\documentclass[letterpaper,twocolumn,10pt]{article}
\usepackage{usenix,epsfig,endnotes}

\usepackage[utf8]{inputenc} 
\usepackage[T1]{fontenc}    
\usepackage{hyperref}       
\usepackage{url}            
\usepackage{booktabs}       
\usepackage{amsfonts}       
\usepackage{nicefrac}       
\usepackage{microtype}      
\usepackage[utf8]{inputenc}
\usepackage{natbib}
\usepackage{amsmath}
\usepackage{amssymb}
\usepackage{amsthm}
\usepackage{mathtools}
\usepackage{subcaption}
\usepackage{caption}
\usepackage[ruled,commentsnumbered]{algorithm2e}
\usepackage[noend]{algpseudocode}
\usepackage{setspace}
\usepackage{graphicx}
\usepackage[margin=1in]{geometry}
\usepackage{hyperref}
\usepackage{color}
\usepackage{import}
\usepackage{hyperref}
\usepackage{graphics, enumitem}
\hypersetup{
  colorlinks=true,      
  linkcolor=blue,       
  citecolor=magenta,    
  filecolor=cyan,       
  urlcolor=red          
}

\newenvironment{denseitemize}{
\begin{itemize}[topsep=2.5pt, partopsep=0pt, leftmargin=1.5em]
  \setlength{\itemsep}{2.5pt}
  \setlength{\parskip}{0pt}
  \setlength{\parsep}{0pt}
}{\end{itemize}}

\title{Accelerating Multi-modal LLM Gaming Performance via Input Prediction and Mishit Correction}

\author{\hspace{0.4cm} Ziyang Lin \hspace{2cm} Zixuan Sun \hspace{2cm} Sanhorn Chen  
\\ ziyang10@illinois.edu \hspace{0.8cm} zixuans8@illinois.edu \hspace{0.8cm} sanhorn2@illinois.edu  
\\ Xiaoyang Chen \hspace{2cm} Roy Zhao
\\ xc52@illinois.edu \hspace{0.8cm} royzhao2@illinois.edu}

\begin{document}

\maketitle

\section{Abstract}
Real-time sequential control agents are often bottlenecked by inference latency: even modest per-step planning delays can destabilize control and degrade reward \cite{kang2025}. We propose a speculation-and-correction framework that adapts the predict-then-verify philosophy of speculative decoding to model-based control with TD-MPC2 \cite{hansen2024tdmpc2scalablerobustworld}. At each step, a pretrained TD-MPC2 world model and latent-space MPC planner produce a short-horizon action queue and predicted latent rollouts; the agent then executes multiple planned actions without replanning. When the next observation arrives, we measure mismatch between the real encoded latent and the queued predicted latent. For small-to-moderate mismatch, a lightweight learned corrector applies a residual update to the speculative action (distilled \cite{hinton2015distillingknowledgeneuralnetwork} offline from a TD-MPC2 replanning teacher); for large mismatch, the system safely falls back to full replanning and clears stale queues. We study both a gated two-tower MLP corrector and a temporal Transformer corrector to handle local errors and systematic drift. On DMC Humanoid-Walk \cite{mujoco}, our method reduces planning inferences from 500 to 282 (43.6\% fewer calls) and improves end-to-end step latency by 25\%, while incurring only a 7.1\% return drop (935 $\rightarrow$ 869). Ablations show that speculative execution without correction is unreliable in longer horizons, confirming that mismatch-aware recycling is essential for robust speedups under tight latency budgets.

\section{Introduction}
Real-time sequential control agents built on Transformer architectures \cite{attention} often face latency problems: every control step must (i) encode rapidly changing sensory inputs and (ii) generate and verify tokens or actions before the system’s next update frame \cite{kang2025}. Existing accelerator methods such as speculative decoding \cite{yaniv2023}, early exiting \cite{tang2023}, and mixture-of-experts (MoE) routing \cite{gavhane2025} are primarily designed to accelerate the inference process after the actual input has been obtained. While these approaches can indeed reduce latency in post-input inference to some extent, both experimental results and practical deployments show that, even with such acceleration, the model’s response speed in real-time scenarios remains insufficient, which directly leads to performance degradation in latency-sensitive sequential control tasks. In this study, we aim to approach this problem through input prediction for speculative execution in fast interactive settings during control or decision making. We propose to build a principled framework that incorporates input prediction, budgeting, and miss-recycling, which we believe would therefore advance both system efficiency and downstream control quality.

Our research frame of reference is motivated by two recent empirical studies. First, speculative decoding demonstrates that by drafting a segment of probable output tokens and verifying them step by step with the large model, it is possible to achieve a 2--3$\times$ speedup without altering the final distribution \cite{yaniv2023}, thereby showing that the ``predict-then-verify'' paradigm can preserve output correctness. Second, early-exit strategies in unified vision--language models reduce computation in both the encoder and decoder by skipping layers based on inter-layer similarity \cite{tang2023, bajpai2025}. Together, these findings suggest that the highest leverage may come not only from predicting future tokens, but also from forecasting upcoming inputs and identifying which experts are worth computing, followed by speculative execution and verification only over this pruned candidate set.

To this end, our project targets three tightly linked areas of study:
\begin{denseitemize}
\item Speculation Space Selection: can we learn a valid distribution over candidates (token/action/vision-patch/expert-route) that maximizes verified-tokens-per-second while preserving control quality?

\item Speculation budgeting: given a strict latency cap, how many branches (width) and how many steps ahead (depth) should we explore to optimize the speed–accuracy trade-off?

\item Repurposing misses: when a speculation is wrong, can we reuse and correct partial state to avoid full recomputation, and can we strategically drop missed predictions to release memory?
\end{denseitemize}

Our focus is on real-time sequential control problems where low latency directly affects task success, such as robotic motion control, competitive gaming, or high-frequency trading. In particular, our experiments are conducted on continuous-control environments from the Mujoco benchmark \cite{mujoco}, where each control step has a tight time constraint and even slight latency may lead to instability or degraded performance.

\section{Motivation}
\paragraph{Latency is not merely a system metric, it is a performance variable.} A recent study, Win Fast or Lose Slow \cite{kang2025}, introduces latency-sensitive benchmarks for real-time sequential control problems (e.g., robotic control, autonomous driving, and high-frequency trading), and shows that reducing decision latency, even though sometimes achieved at the expense of model quality, can raise downstream performance by large margins. In other words, in real-time decision making and other interactive control settings, the agent that acts sooner often performs better even if it is marginally less "accurate." It also demonstrates that adaptive, latency-aware control outperforms fixed policies, which indicates that latency, especially tail latency, is equally necessary as accuracy for Transformer-based agents. 

\begin{figure}
    \centering
    \includegraphics[width=1\linewidth]{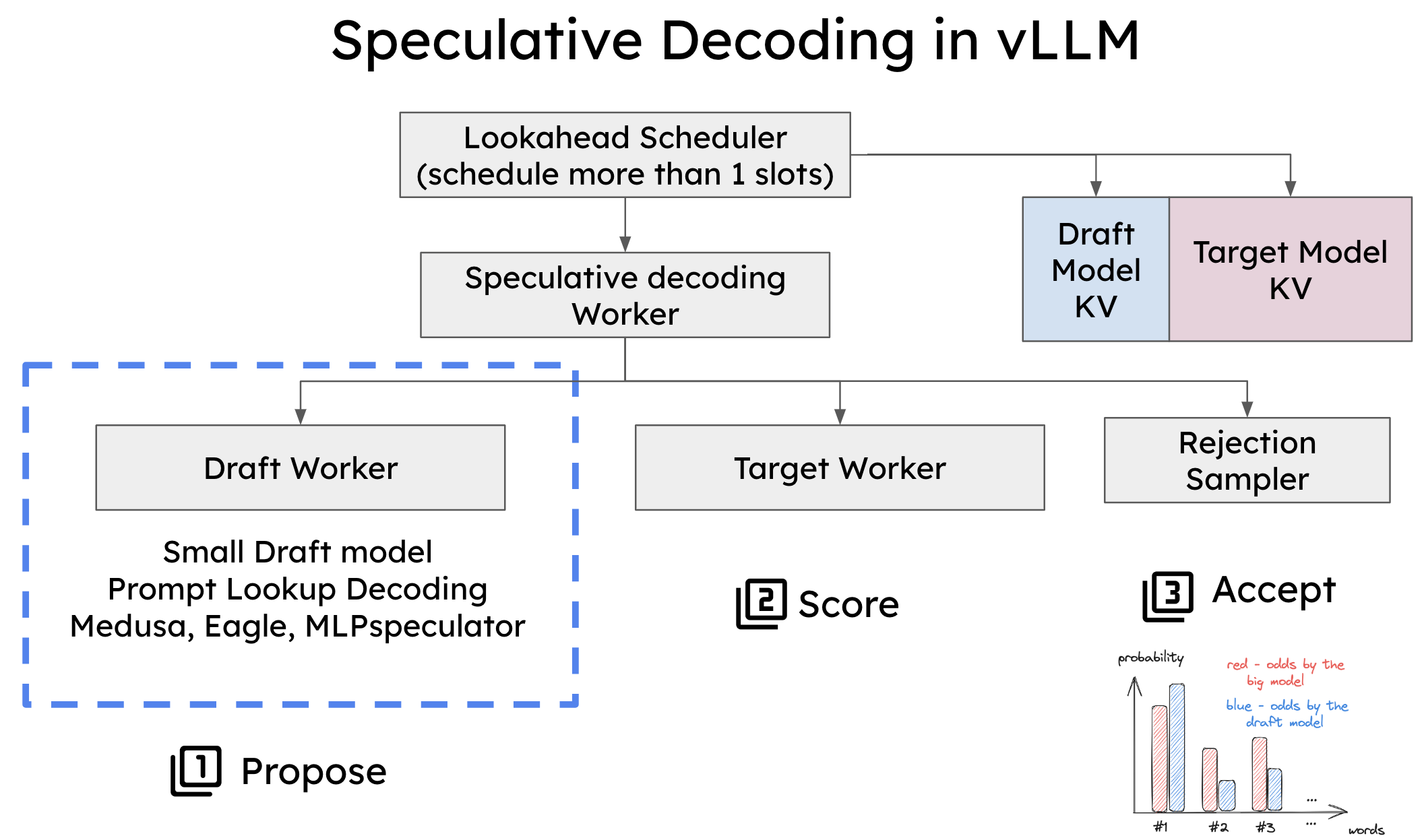}
    \caption{Speculative decoding system diagram (vLLM) \cite{openlm2024} — shows the draft/verify pipeline and KV handling that our approach extends with input prediction and adaptive budgeting.}
    \label{fig:1}
\end{figure}

For Transformer-based control agents, the latency budget is usually split between perception prefill (encoding sensor or visual signals) and autoregressive verification (generating and confirming tokens or actions) \cite{huang2025}. Previous studies offer strong accelerators for each side, but in limited settings. Speculative decoding speeds up the verification process by predicting likely continuations, sometimes just a few more tokens, and verifying them with the main large language model \cite{yaniv2023} as shown in Figure \ref{fig:1}. On the other hand, early-exit methods for Transformers skip encoder and decoder layers based on intermediate similarity, which reveals under-optimized cost in multi-modal or sensor-processing steps \cite{tang2023}. Both approaches indicate two essential principles: predict-then-verify and compute only what you need. However, even with these post-input inference acceleration methods, experimental results show that the model’s response speed remains insufficient in real-time sequential control scenarios, which directly limits overall control quality and task stability \cite{kang2025}.

What's crucial in real-time sequential control is that the next observation is often predictable rather than arbitrary. For example, in robotic manipulation, the next sensor reading strongly depends on the actuator’s previous action and the environment’s dynamics; in high-frequency trading, the next market state is conditioned by recent order flows and self-induced impact. Therefore, the input stream itself is predictable enough to speculate on. If we can learn a compact, high-hit-rate distribution over these futures, forecasting input rather than output, we will be able to convert waiting time into useful precomputation and turn missed forecasts into recyclable features rather than full restarts.

\section{Design}
\begin{figure}[t]
    \centering
    \vspace{1em}
    \includegraphics[width=0.5\textwidth]{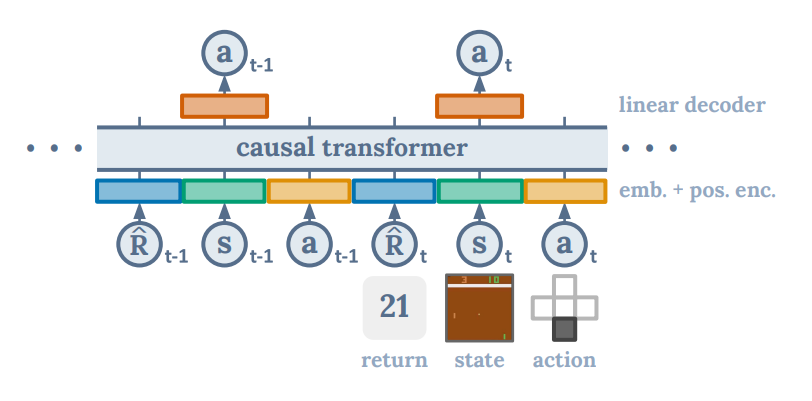}
    \caption{Decision Transformer architecture~\cite{chen2021decisiontransformer}. Each trajectory is represented as an interleaved token sequence of return, state, and action, which are embedded and passed into a causal Transformer to autoregressively predict future actions.}
    \label{fig:dt-arch}
\end{figure}

Transformers, originally designed for text modeling, are inherently \textit{autoregressive}: each generation step depends on the preceding tokens. Consequently, every inference requires one \textit{prefill} pass over the context and several \textit{decode} passes to generate outputs. This sequential dependency makes Transformers naturally less latency-efficient compared to non-autoregressive models, posing a significant limitation in latency-sensitive sequential control domains.

\paragraph{Decision Transformer as a Sequential Control Model.}
Decision Transformer (DT)~\cite{chen2021decisiontransformer} innovatively reinterprets reinforcement learning (RL) as a sequence modeling problem, effectively bridging the gap between autoregressive architectures and real-time control tasks. Instead of predicting the next token in a text, DT models trajectories of states, actions, and returns-to-go as token sequences:
\begin{equation}
\tau = ( \hat{R}_1, s_1, a_1, \hat{R}_2, s_2, a_2, \dots, \hat{R}_T, s_T, a_T ),
\end{equation}
where $s_t$ is the state, $a_t$ the action, and $\hat{R}_t = \sum_{t' = t}^T r_{t'}$ the return-to-go at timestep $t$. During inference, DT conditions on the desired return and the recent $K$ timesteps to autoregressively predict the next action, eliminating the need for complex bootstrapping or value estimation. Figure~\ref{fig:dt-arch} shows the overall causal Transformer structure, where returns, states, and actions are interleaved and passed through modality-specific embeddings before being processed by the Transformer backbone.

This formulation is particularly advantageous for real-time sequential control tasks such as robotic control, gaming, or high-frequency trading, where agents often start from an initial state and execute control sequences step by step. Since most of these tasks start from scratch, the prefill latency is minimal, and each new action requires only one decoding pass, allowing the system to operate efficiently within strict latency budgets. Moreover, the homogeneous activation structure across Transformer layers enables flexible experimentation with layer-wise \textit{miss recycling} and \textit{activation reuse} strategies in later sections.

\paragraph{Speculative Decoding for Autoregressive Transformers.}
Speculative decoding~\cite{yaniv2023} was originally introduced to accelerate autoregressive language models without changing their output distribution. Instead of running a large target model for every next token, a cheaper draft mechanism proposes a block of candidate tokens in parallel; the target model is then invoked once to verify this block via a form of rejection sampling, accepting the longest prefix that matches its own distribution and discarding the rest. This strategy effectively trades additional parallel computation for reduced wall-clock latency, achieving $2\text{--}3\times$ speedups on large Transformers while preserving exact sampling semantics. Our speculative-control scheme is directly inspired by this idea: we treat the Decision Transformer with a next-state head as the “draft” generator of future $(a_t, \hat{s}_{t+1})$ tokens, and then verify or correct these speculative trajectories against real environment states, extending speculative decoding from language generation to closed-loop sequential decision making.

\paragraph{Residual Policy Learning and Policy Decoration.}
Residual policy learning (RPL) \cite{silver2019} provides a simple and powerful template for improving imperfect controllers by learning an additive correction on top of a fixed base policy. Given a hand-designed or model-predictive controller $\pi_0(s)$, RPL learns a residual function $f_\theta(s)$ such that the executed action is $a = \pi_0(s) + f_\theta(s)$, allowing model-free RL to focus on compensating for model errors, sensor noise, and partial observability rather than solving the task from scratch. More recent work such as Policy Decorator \cite{yuan2024} applies the same principle to large imitation-learned policies, training a lightweight residual policy online to refine actions while preserving the smoothness of the underlying base model. Our \emph{Miss Recycling and Correction} module follows this residual-design philosophy: we keep the Decision Transformer backbone fixed and learn a small corrector network that adjusts its speculative action logits using the mismatch between predicted and real states, effectively “decorating’’ the DT with a cache-aware residual policy specialized for speculative execution.

\section{Implementation}
Prior work has attempted to extend the Decision Transformer (DT) architecture by introducing additional prediction heads for the next-state variable \cite{luu2024pcdt, kim2022dadt, wu2023elasticdt}, primarily to enhance generalization rather than to reduce latency. For example, Kim et al.\ proposed the \emph{Dynamics-Augmented Decision Transformer (DADT)}~\cite{kim2022dadt}, which augments the standard return-conditioned DT with an auxiliary head predicting the next state. Their objective was to improve \textit{offline dynamics generalization} by explicitly encoding dynamics information, allowing the model to adapt to unseen transition functions while preserving performance in continuous-control domains. Similarly, Wu et al.\ introduced the \emph{Elastic Decision Transformer (EDT)}~\cite{wu2023elasticdt}, which enables trajectory stitching by dynamically adjusting history length during inference to combine optimal sub-sequences from multiple trajectories. Although both models extend DT with new predictive capabilities, neither focuses on speculative execution to mitigate latency.

\subsection{Speculation Mechanism}
\begin{figure*}[t]
    \centering
    \vspace{1em}
    \includegraphics[width=1\textwidth]{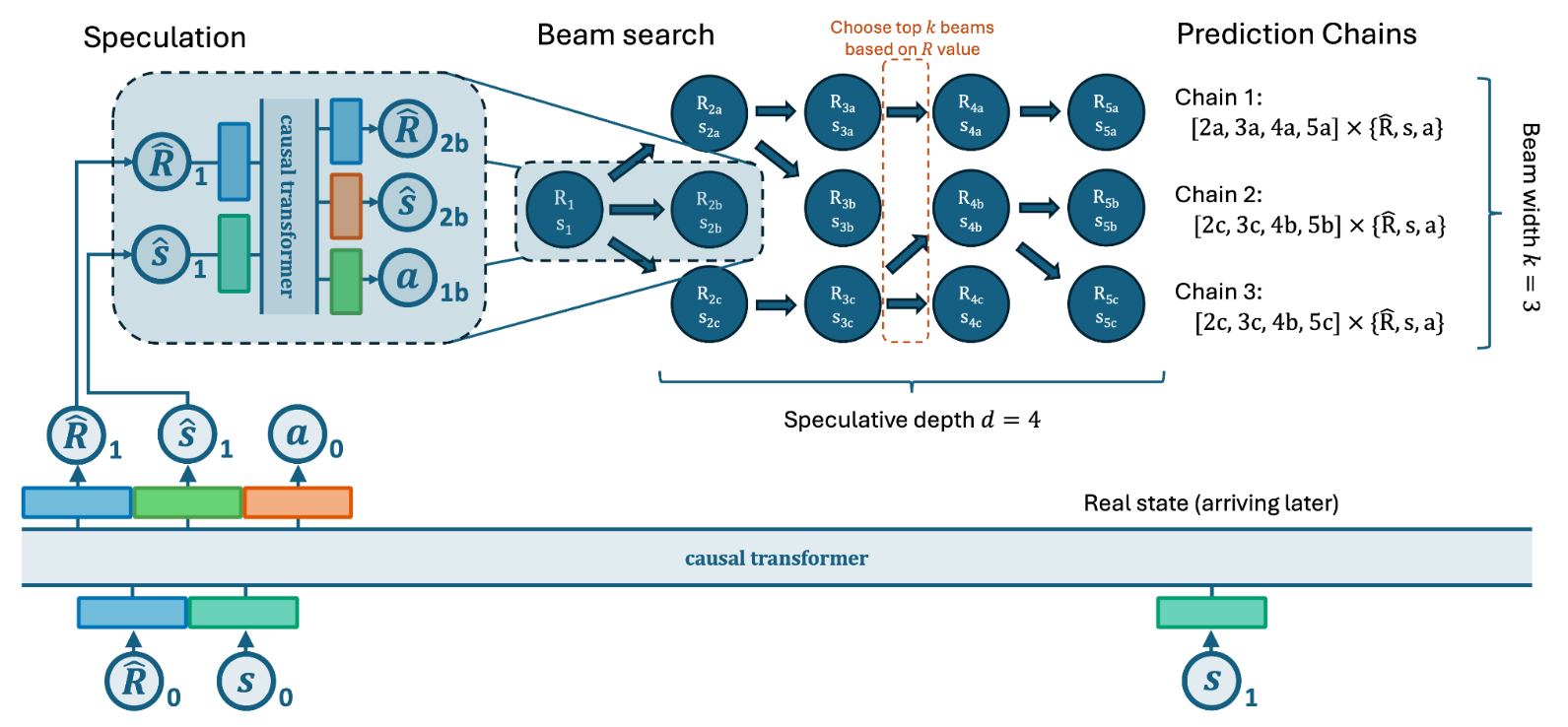}
    \caption{Speculative execution for TD-MPC2. The agent plans a short-horizon action sequence in latent space, executes multiple steps directly, and uses a learned corrector when the real latent deviates from the predicted latent.}
    \label{fig:3}
\end{figure*}

In this work, we adapt TD-MPC2's latent-space model predictive control (MPC) planner \cite{hansen2024tdmpc2scalablerobustworld} into a \emph{speculative execution} mechanism that reduces per-step planning cost. TD-MPC2 naturally produces a short-horizon plan: given the current observation $s_t$, an encoder $E(\cdot)$ maps it into a latent state $z_t = E(s_t)$, and the MPC planner returns an action sequence $\mathbf{a}_{t:t+H-1} = (a_t, \dots, a_{t+H-1})$ together with predicted latent rollouts $\hat{\mathbf{z}}_{t:t+H} = (\hat{z}_t, \hat{z}_{t+1}, \dots, \hat{z}_{t+H})$ under the learned dynamics model $\hat{z}_{\tau+1} = f(\hat{z}_\tau, a_\tau)$.

Rather than replanning at every environment step, we \emph{execute multiple planned steps directly}. Concretely, after computing a plan of horizon $H{=}3$ at time $t$, we execute the subsequent planned actions at times $t{+}1$ and $t{+}2$ without invoking the full MPC again. When the true next state arrives, we encode it as $z^{\text{real}}_{t+1} = E(s_{t+1})$ and compare it to the plan's predicted latent $\hat{z}_{t+1}$. If the mismatch is small, we continue executing the speculative plan; if it is large, we invoke the \emph{Corrector} module to approximate the action that TD-MPC2 would have produced under the real state.

To stress-test long-horizon speculative execution, we further extend the speculation mechanism beyond the default horizon by chaining speculative plans. For example, to execute $L{=}6$ steps while planning with $H{=}3$, we precompute the next block's plan by running the MPC planner starting from the \emph{predicted} latent $\hat{z}_{t+3}$, producing actions for steps $t{+}3$ to $t{+}5$. At runtime, these actions are corrected on-the-fly using the real observations, allowing us to study whether the corrector can prevent error accumulation when speculation spans multiple blocks.

\begin{algorithm}[t]
\caption{Speculative Execution for TD-MPC2 with Corrector}
\label{alg:speculative-tdmpc2}
\footnotesize
\SetAlgoNlRelativeSize{-1}
\DontPrintSemicolon
\SetKwInOut{Require}{Require}
\Require Encoder $E$, dynamics $f$, MPC planner $\text{MPC}(\cdot)$, env $\mathcal{E}$\;
\Require Plan horizon $H{=}3$, execute horizon $L\in\{3,6\}$, threshold $\tau$\;
\Require Corrector $C_\phi$ (optional), history length $K$ (temporal)\;

Initialize $s_0$, action queue $\mathcal{Q}_a \gets \emptyset$, latent queue $\mathcal{Q}_z \gets \emptyset$, history buffer $\mathcal{H}\gets\emptyset$\;

\While{not done}{
  $z^{\text{real}}_t \gets E(s_t)$\;

  \If(\tcp*[f]{Fill queue until we can execute $L$ steps}){$|\mathcal{Q}_a| < L$}{
    \If{$|\mathcal{Q}_z|=0$}{
      $z_{\text{seed}} \gets z^{\text{real}}_t$\;
    }\Else{
      $z_{\text{seed}} \gets \mathcal{Q}_z[\text{end}]$\;
    }
    $(\mathbf{a},\hat{\mathbf{z}})\gets \text{MPC}(z_{\text{seed}},H)$\;
    append $\mathbf{a}$ to $\mathcal{Q}_a$; append $\hat{\mathbf{z}}[1{:}]$ to $\mathcal{Q}_z$\;
  }

  $a^{\text{spec}}_t \gets \mathcal{Q}_a[\text{front}]$; pop front of $\mathcal{Q}_a$\;
  $\hat{z}_t \gets \mathcal{Q}_z[\text{front}]$; pop front of $\mathcal{Q}_z$\;

  $d_t \gets \|z^{\text{real}}_t-\hat{z}_t\|_2$\;

  \uIf{$d_t>\tau$}{
    $(\mathbf{a}^\star,\_)\gets \text{MPC}(z^{\text{real}}_t,H)$\;
    $a_t \gets \mathbf{a}^\star[0]$\;
    clear $\mathcal{Q}_a,\mathcal{Q}_z$\;
  }\Else{
    \uIf{$C_\phi$ enabled}{
      update $\mathcal{H}$ with $(z^{\text{real}}_t,\hat{z}_t,a^{\text{spec}}_t)$ (keep last $K$)\;
      $a_t \gets C_\phi(z^{\text{real}}_t,\hat{z}_t,a^{\text{spec}}_t,\mathcal{H})$\;
    }\Else{
      $a_t \gets a^{\text{spec}}_t$\;
    }
  }

  execute $a_t$ in $\mathcal{E}$ and receive $s_{t+1}$\;
  $t\gets t+1$\;
}
\end{algorithm}

\subsection{Miss Recycling and Correction}

When executing a speculative TD-MPC2 plan, the predicted latent state $\hat{z}_t$ associated with the queued action $a^{\text{spec}}_t$ may deviate from the real latent state $z^{\text{real}}_t = E(s_t)$. A naive response is to discard the speculative queue and re-run the full MPC planner at every deviation, i.e., computing $a_t = \text{MPC}(z^{\text{real}}_t,H)[0]$. While this restores the teacher behavior, it defeats the purpose of speculative execution by incurring an expensive planning call at each mismatch. Instead, we introduce a lightweight \emph{Miss (Mismatch) Recycling and Correction} mechanism that \emph{recycles} the speculative action and predicted latent already computed, and applies a learned corrector network to approximate the action that TD-MPC2 would have produced if it replanned from the real state. Our method is conceptually related to residual policies that refine a frozen base controller~\cite{silver2019,hu2023}, but here the residual is explicitly conditioned on prediction error induced by speculative execution.

At time $t$, Algorithm~\ref{alg:speculative-tdmpc2} provides a speculative pair $(a^{\text{spec}}_t, \hat{z}_t)$ popped from the queues $\mathcal{Q}_a$ and $\mathcal{Q}_z$. After observing the true environment state $s_t$, we compute the real latent $z^{\text{real}}_t = E(s_t)$ and the mismatch
\begin{equation}
    \delta z_t = z^{\text{real}}_t - \hat{z}_t, \qquad d_t = \|\delta z_t\|_2.
\end{equation}
If $d_t$ exceeds the threshold $\tau$, we enter a safe fallback mode and immediately replan with TD-MPC2, discarding the stale speculative queues. Otherwise, for small-to-moderate mismatch we invoke a corrector $C_\phi$ that predicts a residual action update:
\begin{equation}
    \Delta a_t = C_\phi\!\left(z^{\text{real}}_t, \hat{z}_t, a^{\text{spec}}_t, \mathcal{H}_t\right),
\end{equation}
where $\mathcal{H}_t$ optionally denotes a short history of recent mismatch features (used by temporal correctors). The executed action is then
\begin{equation}
    a_t^{\text{corr}} = \mathrm{clip}\!\left(a^{\text{spec}}_t + \Delta a_t\right),
\end{equation}
with $\mathrm{clip}(\cdot)$ enforcing action bounds. The correction requires only a single forward pass through the small corrector network and avoids re-running the MPC planner in the common case, thereby \emph{recycling} speculative computation rather than discarding it.

\subsection{Corrector Architectures}
\begin{figure}[t]
    \centering
    \includegraphics[width=0.95\columnwidth]{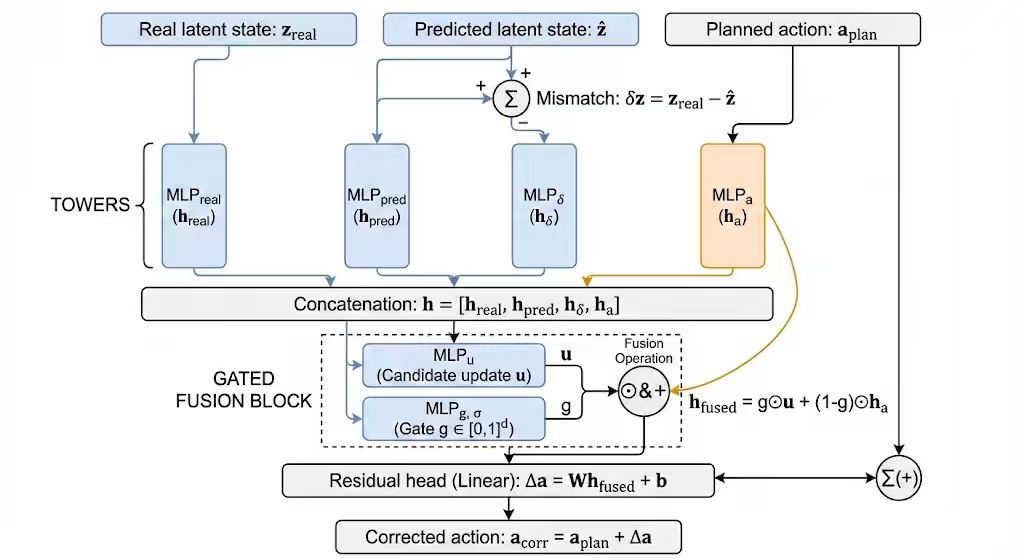}
    \caption{Gated two-tower corrector. Separate towers encode $z^{\text{real}}$ and $\hat{z}$; a mismatch pathway and gate $g$ modulate the residual action update $\Delta a$ applied to the speculative action.}
    \label{fig:gated-mlp}
\end{figure}

\begin{figure}[t]
    \centering
    \includegraphics[width=0.95\columnwidth]{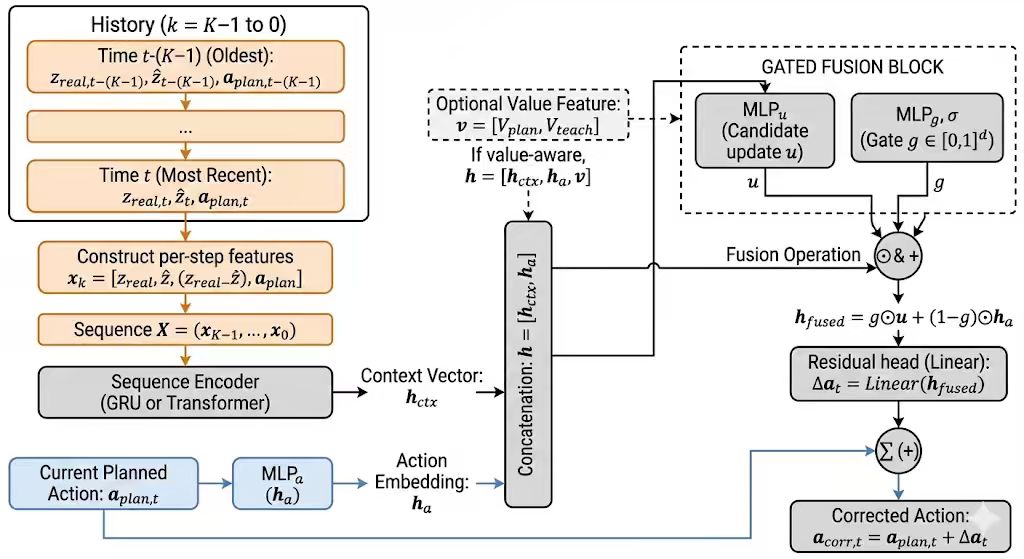}
    \caption{Temporal Transformer corrector. A length-$K$ history of mismatch features is embedded and processed by a Transformer encoder to capture drift over time, producing a residual action correction.}
    \label{fig:temporal-corrector}
\end{figure}

We study two corrector designs to capture different patterns. \emph{Gated two-tower corrector} uses separate feature extractors for the real and predicted latents and fuses them through an explicit mismatch pathway and a learnable gate (see Figure~\ref{fig:gated-mlp}. Concretely, the model forms representations of $z^{\text{real}}_t$, $\hat{z}_t$, and $\delta z_t$ and combines them with an embedding of $a^{\text{spec}}_t$; a gate $g\in[0,1]$ modulates the magnitude of the residual, allowing the corrector to remain near-identity when mismatch is negligible while applying stronger corrections when the mismatch is informative. On the other hand, \emph{Temporal Transformer corrector} conditions on a window of the past $K$ mismatch features (e.g., concatenations of $(z^{\text{real}}, \hat{z}, z^{\text{real}}{-}\hat{z}, a^{\text{spec}})$ over recent steps) and uses self-attention to capture systematic drift and error accumulation across multiple speculative steps (see Figure~\ref{fig:temporal-corrector}). We evaluate both architectures because they represent complementary inductive biases: the two-tower model is fast and reactive for local corrections, while the temporal Transformer is more expressive and is expected to better preserve performance when speculative execution is extended to longer horizons.

We train $C_\phi$ offline by distilling \cite{hinton2015distillingknowledgeneuralnetwork} the TD-MPC2 teacher. During data collection, we run speculative execution (with queued $(a^{\text{spec}},\hat{z})$ pairs) and, at each timestep where mismatch is within the correction regime, we compute the teacher target action by replanning from the real latent:
\begin{equation}
    a_t^\star = \text{MPC}(z^{\text{real}}_t, H)[0].
\end{equation}
The corrector is optimized to match the teacher action while regularizing the residual magnitude:
\begin{equation}
    \mathcal{L}_{\text{corr}} = \big\|a_t^{\text{corr}} - a_t^\star\big\|_2^2 + \lambda \big\|\Delta a_t\big\|_2^2.
\end{equation}
For the temporal Transformer, $\mathcal{H}_t$ is populated with the last $K$ mismatch features (padded at episode starts). To evaluate long-horizon speculative execution, we optionally train with short unrolls (e.g., 2--5 steps) and include multi-step imitation losses to reduce compounding error when the corrector is applied repeatedly between replans.

At test time, the \textit{Miss Recycling and Correction} step in Algorithm~\ref{alg:speculative-tdmpc2} applies $C_\phi$ whenever $d_t \le \tau$, executes $a_t^{\text{corr}}$, and continues consuming the speculative queue; when $d_t > \tau$, the agent safely falls back to a fresh TD-MPC2 replan and clears the speculative buffers. This design combines the latency benefits of speculative multi-step execution with robustness to state prediction mismatch.

\section{Evaluation}
Due to limited time and computational resources, we do not retrain speculative mechanisms directly on top of the Decision Transformer baselines. Instead, we adopt the pretrained TD-MPC2 world model \cite{hansen2024tdmpc2scalablerobustworld} as a standalone speculation module. TD-MPC2 predicts a horizon of three future latent states at each environment step, which we treat as speculative rollouts. These predicted states are combined with a learned corrector to determine how many speculative actions can be safely executed before triggering the next inference.

\paragraph{Experimental Setup.}
\begin{figure}[h] 
\centering \includegraphics[width=0.85\linewidth]{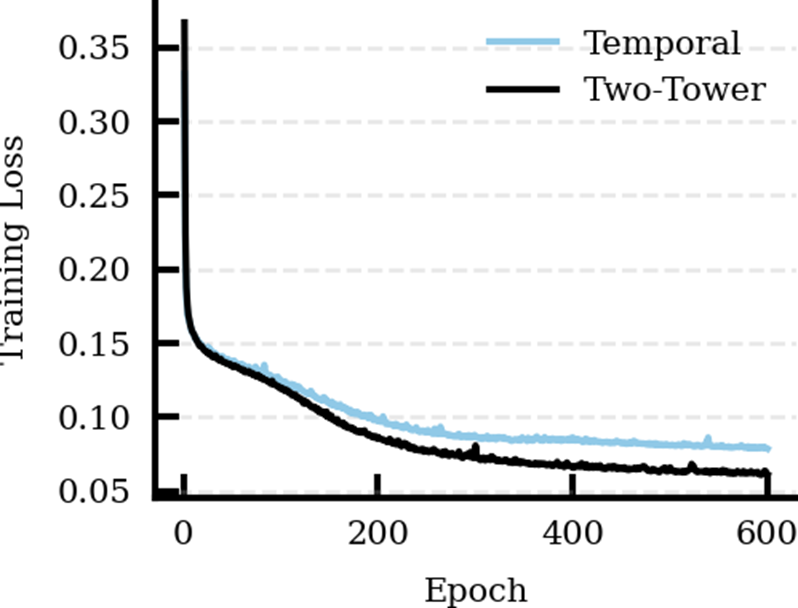} 
\caption{Training loss of the Two-Tower MLP corrector and the Temporal Transformer corrector over 600 epochs on the DMC 
\texttt{Humanoid-Walk} trajectories.} 
\label{fig:training-curves} \end{figure}

We evaluate our approach on the DMC \texttt{Humanoid-Walk} task. Using the TD-MPC2 world model and the main policy, we first collect complete trajectories consisting of ground-truth states, speculative predictions, and executed actions. Based on these trajectories, we train two variants of correctors: a Two-Tower MLP corrector and a Temporal Transformer corrector. Both correctors are trained for 600 epochs until convergence. Figure~\ref{fig:training-curves} shows the training loss curves, where the Temporal corrector converges more smoothly while the Two-Tower model achieves slightly lower final loss.
\begin{figure}[h]
    \centering
    \includegraphics[width=1.05\linewidth]{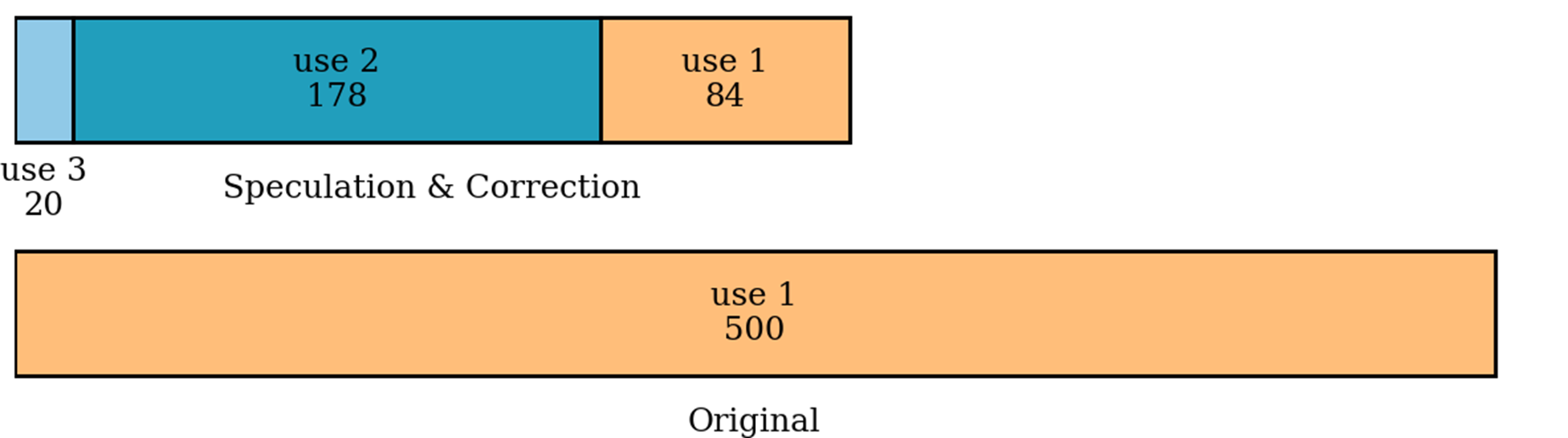}
    \caption{Inference utilization comparison between the original TD-MPC2 execution and the proposed speculation and correction framework over 500 environment steps.}

    \label{fig:infer}
\end{figure}
\paragraph{Inference Reduction via Speculation and Correction.}
We evaluate all methods over 500 environment steps. The original TD-MPC2 baseline performs one inference per environment step, resulting in 500 inference calls. In contrast, our method performs only 282 inference calls in total. Among these, 20 inferences successfully execute all three speculative actions, while 178 inferences execute two speculative actions after correction. This results in a 43.6\% reduction in inference calls compared to the original method. The distribution of executed speculative actions is shown in Figure~\ref{fig:infer}. During evaluation, the majority of inference steps execute exactly two actions, indicating that short-horizon speculation is already highly reliable.
\begin{figure}[h]
    \centering
    \includegraphics[width=0.85\linewidth]{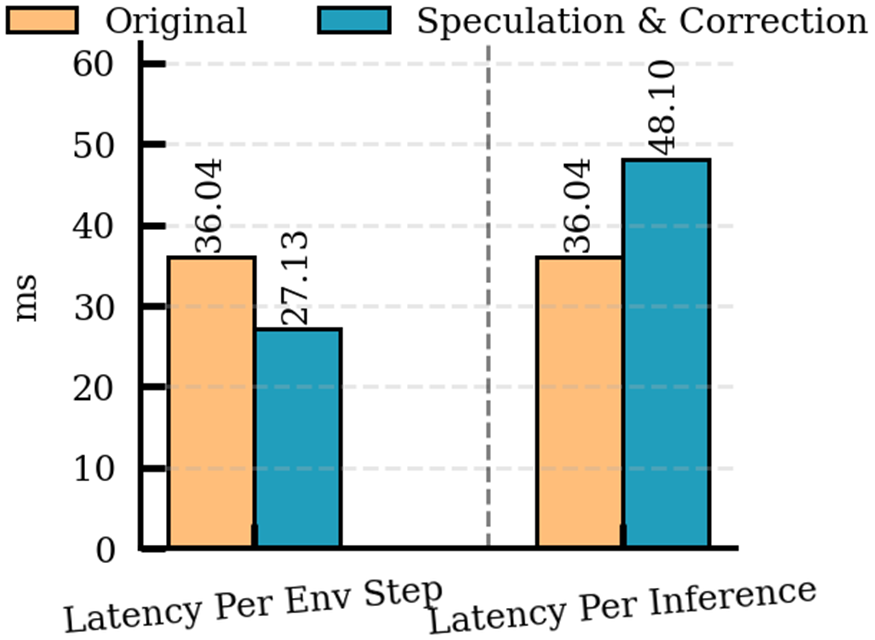}
    \caption{Latency comparison between the original TD-MPC2 method and the proposed speculation and correction approach.}

    \label{fig:latency}
\end{figure}
\paragraph{Latency Analysis.}
We measure both inference-level latency and end-to-end latency per environment step, as shown in Figure~\ref{fig:latency}. The corrector introduces an additional 12~ms overhead per inference due to extra computation. However, because the total number of inferences is substantially reduced, the overall latency per environment step decreases by 9~ms on average. This corresponds to a 25\% speedup compared to the original TD-MPC2 execution, demonstrating that speculation and correction effectively trade per-inference overhead for reduced inference frequency.
\begin{figure}[h]
    \centering
    \includegraphics[width=0.85\linewidth]{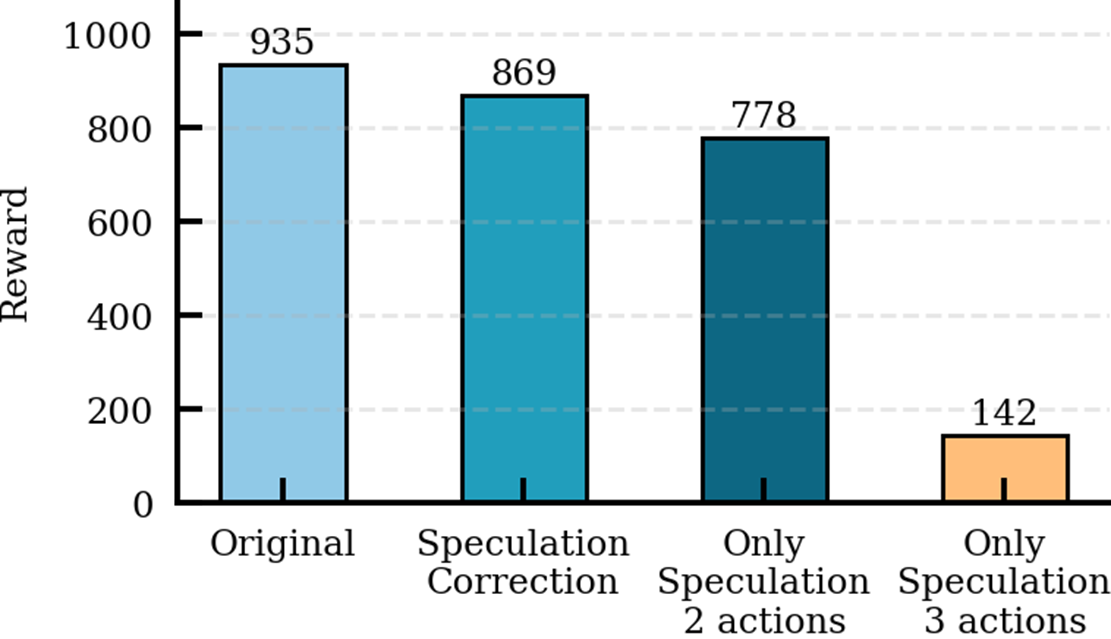}
    \caption{Cumulative reward on the DMC \texttt{Humanoid-Walk} task.}

    \label{fig:reward}
\end{figure}
\paragraph{Reward Performance.}
Figure~\ref{fig:reward} reports the cumulative reward on \texttt{Humanoid-Walk}. The original TD-MPC2 achieves a score of 935, while our method achieves 869, resulting in a drop of only 66 points (approximately 7.1\%). This shows that the latency reduction is achieved with minimal degradation in control performance.

\paragraph{Ablation Studies.}
We further conduct ablation experiments by disabling the corrector and directly executing speculative action chains. Executing three speculative actions without correction leads to a severe performance drop of 793 points (84.8\%), indicating that aggressive speculation without validation is highly unreliable. Executing two speculative actions without correction reduces the reward by 157 points (16.8\%). Compared to this baseline, our full method improves the reward by 91 points, demonstrating that the corrector provides substantial performance gains even when speculation depth is limited.

Overall, while the second speculative action predicted by the TD-MPC2 world model is already of relatively high quality, the corrector consistently improves robustness and reward. These results confirm that speculation combined with lightweight correction offers a favorable trade-off between inference efficiency and control performance in real-time continuous control tasks.

\section{Conclusion}
In this work, we study inference latency as a first-class performance variable in real-time sequential control and propose a speculation-and-correction framework that shifts acceleration from post-input inference to input-side prediction. By leveraging a pretrained world model to speculate future states and a lightweight corrector to recycle missed predictions, our approach significantly reduces inference frequency while preserving control performance. Experiments on the DMC Humanoid-Walk task show that our method achieves a 43.6\% reduction in inference calls and a 25\% end-to-end latency speedup, with only a minor drop in cumulative reward. Ablation results further demonstrate that speculative execution without correction is unreliable, whereas the corrector provides substantial robustness and performance recovery. Overall, our results indicate that combining short-horizon input prediction with lightweight miss correction is a practical and effective strategy for accelerating Transformer-based control agents under strict real-time constraints.

It must be noted that our current evaluation is restricted to a single continuous-control benchmark (Humanoid-Walk) with a fixed planning horizon, so it remains unclear how well the same thresholds and speculation depth transfer to other dynamics regimes or observation modalities. In addition, due to limited time and computational resources, we treat the pretrained TD-MPC2 world model as a standalone speculation module rather than retraining or jointly optimizing the speculation mechanism end-to-end with the downstream policy, which may cap achievable speculation depth before mismatch accumulates. Finally, while the corrector reduces overall per-step latency by reducing the number of MPC calls, it introduces additional per-inference overhead; under harsher real-time budgets or on different hardware, this overhead could narrow the speedup margin.

A natural next step is to jointly train (or online-adapt) the corrector and speculation components to be robust under the exact closed-loop distribution induced by speculative execution, potentially enabling longer horizons with less frequent fallbacks. We also plan to replace the fixed mismatch threshold with an adaptive budgeting rule that selects speculation depth and replanning frequency based on mismatch history and model uncertainty, explicitly optimizing the speed--reward trade-off. Finally, we will broaden evaluation across more DMC tasks and latency-sensitive settings (including noisy sensors and real-world control loops), and explore systems-level pipelining to better overlap speculation, correction, and environment stepping to reduce tail latency.

\section{Team Member Contributions}
\begin{itemize}
\item \emph{ziyang10}: determined the use of the Decision Transformer model, set up the experimental environment, implemented and trained both GPT- and Qwen3-based Decision Transformer models, proposed the speculation solution (Algorithm~1), coordinated the overall team progress, and delivered part of the final project presentation.
\item \emph{zixuans8}: contributed to the design of the speculation-and-correction architecture, formulated the correction module mathematically, defined its technical implementation roadmap, provided the GPU training cluster, and delivered part of the final project presentation.
\item \emph{sanhorn2}: proposed and evaluated an activation-level modification scheme (not included in the final version), explored alternative correction mechanisms, identified appropriate training tasks and experimental benchmarks, and delivered part of the final project presentation.
\item \emph{xc52}: assisted in building the experimental environment, supported code implementation and validation, provided a local lightweight testing platform, and prepared presentation slides for the final project presentation.
\item \emph{royzhao2}: conducted literature review, prepared experimental visualizations, assisted in writing the experimental report, and prepared presentation slides for the final project presentation.
\end{itemize}

\bibliographystyle{unsrt}
\bibliography{rlps}

@misc{kang2025,
      title={Win Fast or Lose Slow: Balancing Speed and Accuracy in Latency-Sensitive Decisions of LLMs}, 
      author={Hao Kang and Qingru Zhang and Han Cai and Weiyuan Xu and Tushar Krishna and Yilun Du and Tsachy Weissman},
      year={2025},
      eprint={2505.19481},
      archivePrefix={arXiv},
      primaryClass={cs.LG},
      url={https://arxiv.org/abs/2505.19481}, 
}

@misc{hinton2015distillingknowledgeneuralnetwork,
      title={Distilling the Knowledge in a Neural Network}, 
      author={Geoffrey Hinton and Oriol Vinyals and Jeff Dean},
      year={2015},
      eprint={1503.02531},
      archivePrefix={arXiv},
      primaryClass={stat.ML},
      url={https://arxiv.org/abs/1503.02531}, 
}

@INPROCEEDINGS{tang2023,
  author={Tang, Shengkun and Wang, Yaqing and Kong, Zhenglun and Zhang, Tianchi and Li, Yao and Ding, Caiwen and Wang, Yanzhi and Liang, Yi and Xu, Dongkuan},
  booktitle={2023 IEEE/CVF Conference on Computer Vision and Pattern Recognition (CVPR)}, 
  title={You Need Multiple Exiting: Dynamic Early Exiting for Accelerating Unified Vision Language Model}, 
  year={2023},
  volume={},
  number={},
  pages={10781-10791},
  doi={10.1109/CVPR52729.2023.01038}}

@misc{gavhane2025,
      title={MoE-Beyond: Learning-Based Expert Activation Prediction on Edge Devices}, 
      author={Nishant Gavhane and Arush Mehrotra and Rohit Chawla and Peter Proenca},
      year={2025},
      eprint={2508.17137},
      archivePrefix={arXiv},
      primaryClass={cs.LG},
      url={https://arxiv.org/abs/2508.17137}, 
}

@inproceedings{yaniv2023,
author = {Leviathan, Yaniv and Kalman, Matan and Matias, Yossi},
title = {Fast inference from transformers via speculative decoding},
year = {2023},
publisher = {JMLR.org},
booktitle = {Proceedings of the 40th International Conference on Machine Learning},
articleno = {795},
numpages = {13},
location = {Honolulu, Hawaii, USA},
series = {ICML'23}
}

@misc{bajpai2025,
      title={FREE: Fast and Robust Vision Language Models with Early Exits}, 
      author={Divya Jyoti Bajpai and Manjesh Kumar Hanawal},
      year={2025},
      eprint={2506.06884},
      archivePrefix={arXiv},
      primaryClass={cs.LG},
      url={https://arxiv.org/abs/2506.06884}, 
}

@misc{huang2025,
      title={SpecVLM: Fast Speculative Decoding in Vision-Language Models}, 
      author={Haiduo Huang and Fuwei Yang and Zhenhua Liu and Xuanwu Yin and Dong Li and Pengju Ren and Emad Barsoum},
      year={2025},
      eprint={2509.11815},
      archivePrefix={arXiv},
      primaryClass={cs.CV},
      url={https://arxiv.org/abs/2509.11815}, 
}

@misc{openlm2024,
  title   = {Speculative Decoding in vLLM},
  author  = {{OpenLM}},
  year    = {2024},
  month   = {October},
  url     = {https://openlm.ai/speculative-decoding-in-vllm/},
  note    = {Diagram illustrating how draft and target runners interact within the vLLM batching system}
}

@inproceedings{chen2021decisiontransformer,
  title        = {Decision Transformer: Reinforcement Learning via Sequence Modeling},
  author       = {Lili Chen and Kevin Lu and Aravind Rajeswaran and Kimin Lee and Aditya Grover and Michael Laskin and Pieter Abbeel and Aravind Srinivas and Igor Mordatch},
  booktitle    = {Proceedings of the 35th International Conference on Neural Information Processing Systems (NeurIPS 2021)},
  series       = {NeurIPS '21},
  pages        = {15084--15097},
  year         = {2021},
  publisher    = {Curran Associates, Inc.},
  url          = {https://proceedings.neurips.cc/paper/2021/hash/7f489f642a0ddb10272b5c31057f0663-Abstract.html},
  doi          = {10.5555/3504035.3504563}
}

@inproceedings{kim2022dadt,
  title        = {Dynamics-Augmented Decision Transformer for Offline Dynamics Generalization},
  author       = {Changyeon Kim and Junsu Kim and Younggyo Seo and Kimin Lee and Honglak Lee and Jinwoo Shin},
  booktitle    = {3rd Offline Reinforcement Learning Workshop at NeurIPS 2022: Offline RL as a “Launchpad”},
  year         = {2022},
  note         = {Poster/Workshop paper; OpenReview link https://openreview.net/pdf?id=ReNyLYfUdr},
  url          = {https://openreview.net/pdf?id=ReNyLYfUdr}
}

@inproceedings{wu2023elasticdt,
  title        = {Elastic Decision Transformer},
  author       = {Yueh-Hua Wu and Xiaolong Wang and Masashi Hamaya},
  booktitle    = {Proceedings of the 37th International Conference on Machine Learning (ICML) – Workshop/NeurIPS 2023 Poster},
  series       = {NeurIPS '23},
  pages        = {–}, 
  year         = {2023},
  publisher    = {Curran Associates, Inc.},
  doi          = {10.5555/3666122.3666936},
  url          = {https://dl.acm.org/doi/10.5555/3666122.3666936}
}

@inproceedings{luu2024pcdt,
  title        = {Predictive Coding for Decision Transformer},
  author       = {Tung M. Luu and Donghoon Lee and Chang D. Yoo},
  booktitle    = {2025 IEEE/RSJ International Conference on Intelligent Robots and Systems (IROS 2025)},
  year         = {2025},
  note         = {Accepted for publication; arXiv preprint arXiv:2410.03408v1},
  url          = {https://arxiv.org/abs/2410.03408v1}
}

@misc{silver2019,
      title={Residual Policy Learning}, 
      author={Tom Silver and Kelsey Allen and Josh Tenenbaum and Leslie Kaelbling},
      year={2019},
      eprint={1812.06298},
      archivePrefix={arXiv},
      primaryClass={cs.RO},
      url={https://arxiv.org/abs/1812.06298}, 
}

@misc{hu2023,
      title={Decision Transformer under Random Frame Dropping}, 
      author={Kaizhe Hu and Ray Chen Zheng and Yang Gao and Huazhe Xu},
      year={2023},
      eprint={2303.03391},
      archivePrefix={arXiv},
      primaryClass={cs.LG},
      url={https://arxiv.org/abs/2303.03391}, 
}

@inproceedings{attention,
 author = {Vaswani, Ashish and Shazeer, Noam and Parmar, Niki and Uszkoreit, Jakob and Jones, Llion and Gomez, Aidan N and Kaiser, \L ukasz and Polosukhin, Illia},
 booktitle = {Advances in Neural Information Processing Systems},
 editor = {I. Guyon and U. Von Luxburg and S. Bengio and H. Wallach and R. Fergus and S. Vishwanathan and R. Garnett},
 pages = {},
 publisher = {Curran Associates, Inc.},
 title = {Attention is All you Need},
 url = {https://proceedings.neurips.cc/paper_files/paper/2017/file/3f5ee243547dee91fbd053c1c4a845aa-Paper.pdf},
 volume = {30},
 year = {2017}
}

@INPROCEEDINGS{mujoco,
  author={Todorov, Emanuel and Erez, Tom and Tassa, Yuval},
  booktitle={2012 IEEE/RSJ International Conference on Intelligent Robots and Systems}, 
  title={MuJoCo: A physics engine for model-based control}, 
  year={2012},
  volume={},
  number={},
  pages={5026-5033},
  doi={10.1109/IROS.2012.6386109}}

@misc{yuan2024,
      title={Policy Decorator: Model-Agnostic Online Refinement for Large Policy Model}, 
      author={Xiu Yuan and Tongzhou Mu and Stone Tao and Yunhao Fang and Mengke Zhang and Hao Su},
      year={2024},
      eprint={2412.13630},
      archivePrefix={arXiv},
      primaryClass={cs.RO},
      url={https://arxiv.org/abs/2412.13630}, 
}

@misc{hansen2024tdmpc2scalablerobustworld,
      title={TD-MPC2: Scalable, Robust World Models for Continuous Control}, 
      author={Nicklas Hansen and Hao Su and Xiaolong Wang},
      year={2024},
      eprint={2310.16828},
      archivePrefix={arXiv},
      primaryClass={cs.LG},
      url={https://arxiv.org/abs/2310.16828}, 
}

\end{document}